\documentclass{article}

\usepackage{microtype}
\usepackage{graphicx}
\usepackage{subfigure}
\usepackage{mathtools}
\usepackage{booktabs} 
\usepackage{amsmath, amssymb, amsthm}
\newcommand{\Dataset}{ \mathcal{D} }
\newcommand{\reals}{ \mathbb{R} }
\usepackage{bm}
\usepackage{hyperref}

\usepackage{xcolor}

\usepackage[accepted]{icml2020}

\icmltitlerunning{Reliable Uncertainties for Bayesian Neural Networks using
Alpha-divergences}

\begin{document}

\twocolumn[
\icmltitle{Reliable Uncertainties for Bayesian Neural Networks\\ using
Alpha-divergences}



\icmlsetsymbol{equal}{*}

\begin{icmlauthorlist}
\icmlauthor{H\'ector Javier Hort\'ua}{equal,to}
\icmlauthor{Luigi Malago}{equal,to}
\icmlauthor{Riccardo Volpi}{equal,to}
\end{icmlauthorlist}

\icmlaffiliation{to}{Machine Learning and Optimization Group,
Romanian Institute of Science and Technology (RIST),
Cluj-Napoca, Romania}

\icmlcorrespondingauthor{H\'ector J. Hort\'ua}{hortua.orjuela@rist.ro}

\icmlkeywords{Machine Learning, ICML}

\vskip 0.3in
]



\printAffiliationsAndNotice{\icmlEqualContribution} 

\begin{abstract}
Bayesian Neural Networks (BNNs) often result uncalibrated after training, usually tending towards overconfidence. Devising effective calibration methods with low impact in terms of computational complexity is thus of central interest.
In this paper we  present calibration methods for BNNs based on the alpha divergences from Information Geometry.
We compare the use of alpha divergence in training and in calibration, and we show how the use in calibration provides better-calibrated uncertainty estimates for specific choices of alpha and is more efficient especially for complex network architectures.
We empirically demonstrate the advantages of alpha calibration in regression problems involving parameter estimation and inferred correlations between output uncertainties.
\end{abstract}

\section{Introduction}
Bayesian neural networks (BNNs)~\cite{10.1145/168304.168306}  have recently received a great deal of interest due to their ability of quantifying the uncertainty in the predicted parameters and provide estimates for the ignorance of the model. However, current algorithms built to estimate uncertainty in deep neural networks (included BNNs) often include overconfidence issues that  makes them  inappropriate to be deployed for real-world problems.  In the literature several techniques have been proposed to calibrate the network after training  such as  Platt, vector-matrix or  Temperature scaling~\cite{Platt99probabilisticoutputs},  and non-parametric ones like Histogram binning~\cite{Zadrozny:2001:OCP:645530.655658},  Isotonic regression~\cite{Zadrozny:2002:TCS:775047.775151}, or  Beta calibration~\cite{pmlr-v54-kull17a} which yields to a well calibrated networks~\cite{Guo:2017:CMN:3305381.3305518}.
All these methods are post-processing steps i.e., applied after training in order to ensure the performance in predictions. Furthermore, existing methods applied during training  are  found in~\cite{hortua2019parameters,PerreaultLevasseur:2017ltk,Gal2016Bayesian} where by tuning hyper-parameters such as dropout rate or L2-regularizers provide accurate uncertainties.
On the other hand, an additional source of miscalibration for  BNNs arises due to the approximate methods used for obtaining the  posterior distribution for the model parameters. In fact, variational inference (VI)~\cite{graves2011practical} or expectation propagation (EP)~\cite{2013arXiv1301.2294M} provide the scenario in which BNNs capture a density close to the exact posterior which carries information about the uncertainty on the predictions~\cite{minka2005divergence}. However, the variational distribution might not  match the exact posterior, leading  to a  worse predictive distribution and  poor  uncertainty estimates. In~\cite{li2017dropout,minka2005divergence,2019arXiv190906945R,2015arXiv151103243H} 
 alternative divergence choices were included during the optimization processes,  resulting  in improved uncertainty estimates and accuracy compared to the traditional VI (adapted only at KL divergence). 
 However,  it is still not clear what is the most convenient and reliable way to calibrate BNNs. Post-processing methods does not calibrate in all cases the approximate posterior obtained by VI and the calibration during training could not achieve the  state-of-the-art of  neural networks.
 This paper aims to fill this gap and analyse the performance of different methods commonly used for calibration in regression tasks. Additionally, we proposed a new method which combine most prominent post processing methods along with information geometry to calibrate BNNs. Applying these methods, we have obtained well-calibrated networks for specific choices of the $\alpha-$hyper-parameter. The proposed methods are straightforward to implement and efficient also for complex architectures.

\section{Background}\label{sectheory}
Given a training dataset $\Dataset=\{({\bm x}_i, {\bm y}_i)\}_{i=1}^D$ formed by $D$ couples of features ${\bm x}_i \in \reals^M $  and their respective targets ${\bm y}_i \in \reals^N$, BNNs are defined through a prior  $p({\bm w})$ on the model parameters ${\bm w}$  and the likelihood of the model $p({\bm y}|{\bm x},{\bm w})$.  Variational Inference (VI) method  allows to  approximate the real posterior by a parametric distribution $q({\bm w}|\theta)$  depending on a set of variational parameters ${\bm \theta}$. These parameters are adjusted to minimize a certain divergence, usually given by the KullBack-Leibler divergence between the true posterior (generally intractable) and the approximate posterior $\text{KL}(q({\bm w}|{\bm \theta})||p({\bm w}|\mathcal{D}))$. It has been shown that minimizing this KL divergence is equivalent to minimizing the following objective function~\cite{NIPS2011_4329}
\begin{eqnarray}\label{eq:4}
\mathcal{L}_{VI} &=& \text{KL}(q({\bm w}|{\bm \theta})||p({\bm w}))\nonumber\\
&-&\sum_{(\bm{x},\bm{y})\in\Dataset}\int_\Omega q({\bf w}|{\bm \theta}) \ln p({\bf y}|{\bm x},{\bm w}) d{\bm w} \; .
\end{eqnarray}
Once  ${{\bm \theta}}$ is learned, we can make predictions via Monte Carlo sampling
\begin{equation}\label{eq:5}
 q_{\hat{{\bm \theta}}}({\bf y}^*|{\bm x}^*) \approx \frac{1}{K}\sum_{k=1}^K p({\bf y}^*|{\bm x}^*,\hat{{\bm w}}_k) \;\;\;\; \mbox{with  }
 \hat{{\bm w}}_k \sim q({\bm w}|{\bm\theta}),
\end{equation}
where $K$ is the number of samples and $^*$ represents a new input data for inference. To infer the correlations between the  parameters for regression tasks~\cite{hortua2019parameters}, we need to predict the full covariance matrix. This requires to produce in output of the last layer of the network a mean vector $\bm{\mu}\in\mathbb{R}^{N}$ and the lower triangular Cholesky decomposition of the covariance matrix $\Sigma\in\mathbb{R}^{N(N+1)/2}$ that represents the aleatoric uncertainty. These will define the parameters of the Multivariate Gaussian distribution output of the model, and the negative log-likelihood $\text{NLL}\sim -\ln p({\bm y}|{\bm x},{\bm w})$ can be computed as~\cite{Dorta_2018,Cobb_2019}
\begin{equation}
\label{eq:16}
 \text{NLL}\sim \frac{1}{2}\log |\Sigma|+ \frac{1}{2}({\bm y}-{\bm \mu})^\top \Sigma^{-1}({\bm y}-{\bm \mu}).
\end{equation}
Recently, extensions to a  more rich family of divergences with the purpose of approximating better the posterior distribution of the weights has been introduced  in~\cite{2015arXiv151103243H} and studied in detail in~\cite{2019arXiv190906945R,li2017dropout}. The Black-box alpha(BB-$\alpha$)  method   relies on  the energy function used by power EP method~\cite{minka2004power} and focuses on  the minimization of the local $\alpha$-divergences defined as
\begin{equation}
\label{eq:alpha_divergence}
\mathrm{D}_{\alpha}[p||q] = \frac{1}{\alpha (1 - \alpha)} \left(1 - \int p(x)^{\alpha} q(x)^{1 - \alpha} dx \right),
\end{equation}
where $\alpha= 0$ is used in VI and $\alpha= 1.0$ is used in EP, while the case $\alpha= 0.5$ is known as Hellinger distance and $\alpha = 2$ is the $\chi^2$ distance.  In the limit of $\alpha/D\rightarrow0$, the authors in~\cite{2015arXiv151103243H,li2017dropout} arrive to a generalization of Eq.~\ref{eq:4}  given by
\begin{align}\label{BBalphalossF}
  \mathcal{L}_\alpha &\approx \text{KL}(q({\bm w}|{\bm \theta})||p({\bm w}))\nonumber\\
  &-\frac{1}{\alpha} \sum_{(\bm{x},\bm{y})\in\Dataset} \ln \int_\Omega q({\bf w}|{\bm \theta})p({\bm y} | {\bm x}, {\bm w})^\alpha d{\bm w} \;,
\end{align}
that allows to optimize a family of divergences depending from a parameter $\alpha$, resulting in approximate distributions $q$ with different properties. We can observe that  fixing $\alpha= 1$, the per-point predictive log-likelihood $ \log \mathbb{E}_{q} \left[p({\bm y}_i | {\bm x}_i, {\bm w}) \right]$ is  directly  optimised; while for $\alpha\rightarrow0$  and sampling once, the Eq.~\ref{BBalphalossF} reduces to the original stochastic VI loss function Eq.~\ref{eq:4}, where the optimization will be focus on $ \mathbb{E}_{q} \left[\log p({\bm y}_i | {\bm x}_i, {\bm w}) \right]$. 

\section{Related Work}
Post-processing calibration techniques can be either parametric or non-parametric~\cite{Guo:2017:CMN:3305381.3305518}. Temperature Scaling (TS) is often the most effective and simple technique that improve the calibration. It has  been extensively  studied in  the  literature and some extensions  have been developed recently~\cite{Guo:2017:CMN:3305381.3305518,pmlr-v54-kull17a,2019arXiv190511659L,kuleshov2018accurate}. 
TS is the simplest extension  of  Platt scaling, and  for regression tasks (studied  in~\cite{2019arXiv190511659L})  consists in  multiplying  the variance of each predicted distribution by a single scalar parameter $s>0$. We will refer to this method as {\bf sTS} through the rest of the paper. The  predicted Gaussian distribution $\mathcal{N}({\bm \mu},{\bm \Sigma})$, is modified as  $\mathcal{N}({\bm \mu},s{\bm \Sigma})$ and the calibration optimizes the objective function with respect to the scalar $s$. The parameters of the network  are fixed during this stage, implying that sTS does not modify the prediction performances since it is not affecting the $\mu$. Hence, this method is good in calibrating aleatoric uncertainties, but might not be suitable for BNNs in which epistemic uncertainty must also be taken into account.  
\section{Proposed calibration methods}\label{secmethods}
We extend the TS method by using the BB-$\alpha$ objective.
Introducing a lower triangular matrix, ${\bm L}$, instead of a single scalar parameter, leads to an anisotropic scaling. The covariance matrix in Eq.~\ref{eq:16} is scaled as $\Sigma\rightarrow L^\top\Sigma L$, ensuring the positive semi definite property. This method (hereafter called {\bf TrilTS})  has the ability to calibrate not only the predicted variance but also the principal directions of variations for the output.
Both the presented methods are still not able to affect the mean output predictions and thus are not able to calibrate epistemic uncertainties in a Bayesian setting.
An alternative is to perform the network calibration by fine-tuning the last output layer with the alpha divergence (after the network has been trained with standard KL).
Additionally, the LL method can be aided in the optimization by additionally using either a scalar ({\bf sLL}) or a lower triangular ({\bf TrilLL}) temperature parameter. We can also consider retraining only the weights related to the inferred means, which ultimately are the one affecting epistemic uncertainty. We will refer to this method {\bf LL$\mu$}.
Again, it can be implemented with the scalar ({\bf sLLmean}) or matrix ({\bf TrilLLmean}) parameter. These methods make trying several values of alpha more efficient since fine-tuning the last layer is much cheaper than re-training the full network, especially for complex architectures.
\section{Experimental Set Up}\label{secimplementation}
{\bf Dataset}: We generated 50.000 images related to the Cosmic Microwave Background (CMB)  maps projected in $20\times20 deg^2$ patches in the sky using the script described in~\cite{hortua2019parameters}. These images have size of (256,256,3) and each image corresponds to a specific set value of three  parameters. They are split in $70\%$ for training, $10\%$ for validation and the rest for testing.\\
{\bf Architecture}: All the networks are implemented using  TensorFlow~\footnote{\MakeLowercase{h}ttps://www.tensorflow.org/} and TensorFlow-Probability~\footnote{\MakeLowercase{h}ttps://www.tensorflow.org/probability}. We used a modified version of the VGG architecture with 5 VGG blocks (each made by two Conv2D layers and one max pooling) and channels size [32, 32, 32, 32, 64]. Kernel size is fixed to 3$\times$3 and activation function used is LeakyReLU ($\alpha=-0.3$)  following by a batch renormalization layer.   We used the Flipout method, assuming in Eq.~\ref{eq:4} a Gaussian distribution over the weights for both prior and posterior, and providing an efficient way to draw pseudo-independent weights for different elements in a single batch~\cite{wen2018flipout}.\\
{\bf Evaluation Metrics}: In order to quantify the performance of the networks, we computed the predicted interval coverage probabilities displayed in the the reliability diagrams (see Appendix~\ref{appedA}), and the  coefficient of determination
\begin{align}
  R^2&=1-\frac{\sum_i (\bm{\bar{\mu}}(\bm{x}_i)-\bm{y}_i)^2}{\sum_i (\bm{y}_i-\bm{\bar{y}})^2},
\end{align}
where $\bm{\bar{\mu}}(\bm{x}_i)$ are the predicted values of the trained BNN, $\bm{\bar{y}}$ is the average of the true parameters and the summations are performed over the entire test set. $R^2$ ranges from 0 to 1, where 1 represents perfect inference. Thus, the coverage probabilities (and the test NLL) allow to evaluating the fidelity of posterior approximations, while $R^2$ measures the accuracy of the regression model.
We will use Beta calibration~\cite{pmlr-v54-kull17a} as a baseline to compare the standard calibration methods with respect to our approaches, hereafter referred  as $\beta$-BNN.
\section{Results}\label{secexp}
\begin{figure}[h!]
  \centering
    \includegraphics[width=0.5\textwidth]{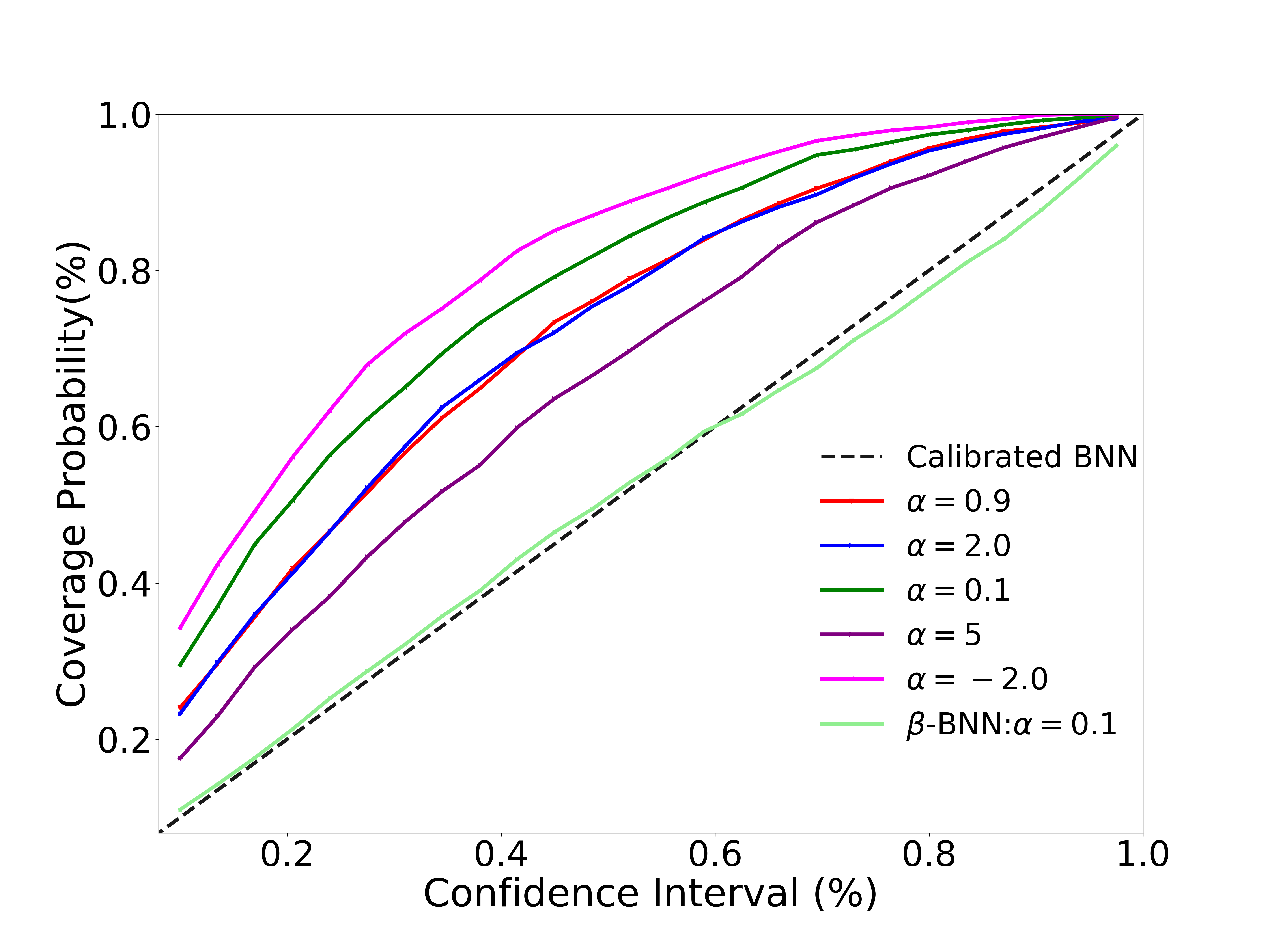}
     \caption{Reliability diagrams for BNNs using different $\alpha$-values  without  taking any post-process calibration technique. }
     \label{reliabtraining}
\end{figure}

We start by evaluating the performance of BB-$\alpha$ on Bayesian neural networks employing Flipout to sample from the Gaussian distributions over the weights~\cite{wen2018flipout}. We consider $\alpha \in [-2, 3]$.  We  summarise  accuracy in the uncertainties through the reliability diagrams displayed in Fig.~\ref{reliabtraining}.
Even extending the range of $\alpha$, we observe that we cannot obtain calibrated BNNs during training. Large $\alpha-$values mitigate the problem (although
we will see that too large values tend to produce numerical
instabilities), but no value of alpha can calibrate the network during training. 

\begin{figure}[h!]
  \centering
    \includegraphics[width=0.5\textwidth]{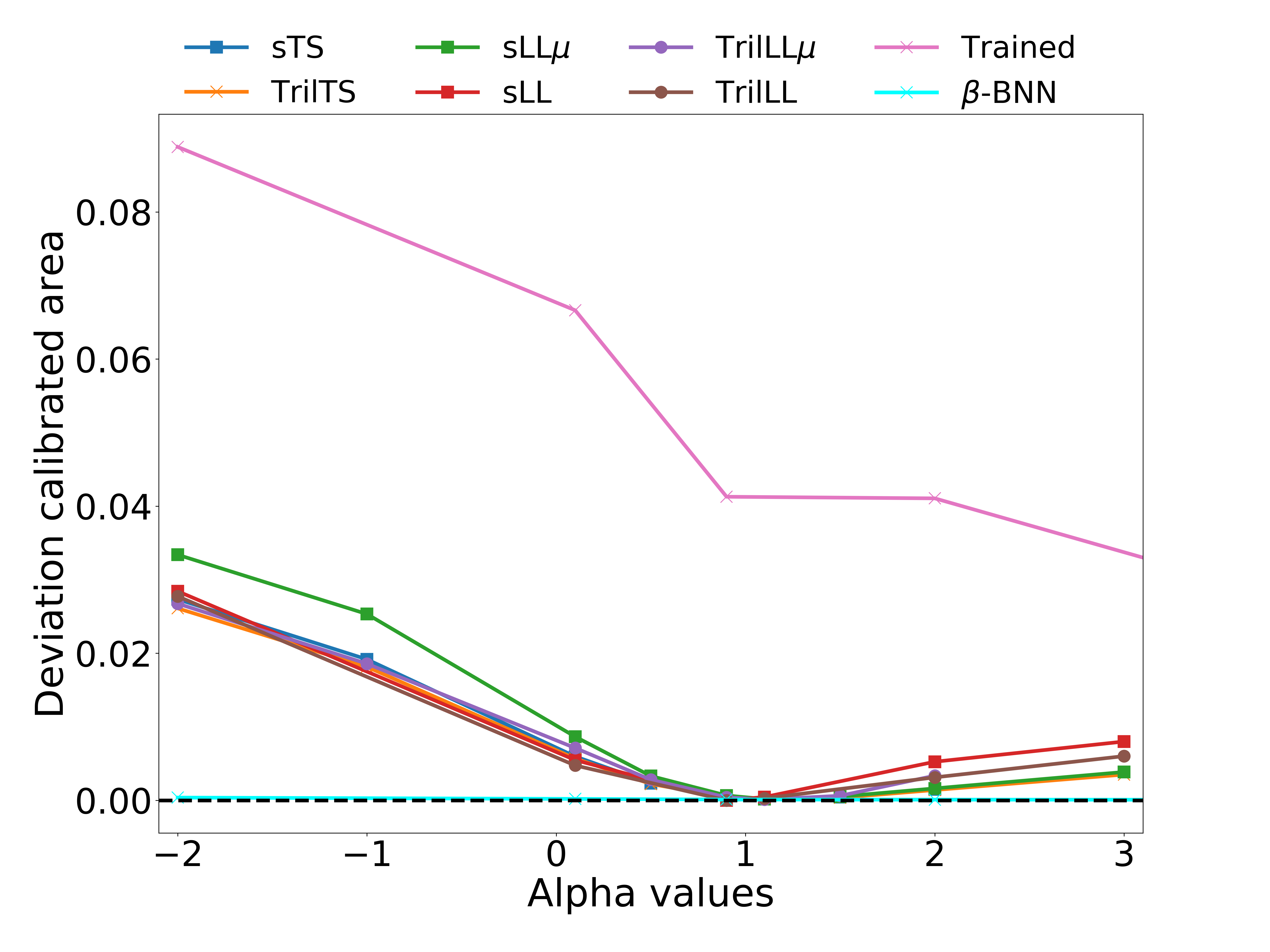}
     \caption{ Area between  different calibration methods and perfectly calibrated networks. The perfect calibration model corresponds to the $\alpha$-values where occur the  intersection of the curves with the solid black line.}
     \label{figarea}
\end{figure}

\begin{figure}[h!]
  \centering
    \includegraphics[width=0.5\textwidth]{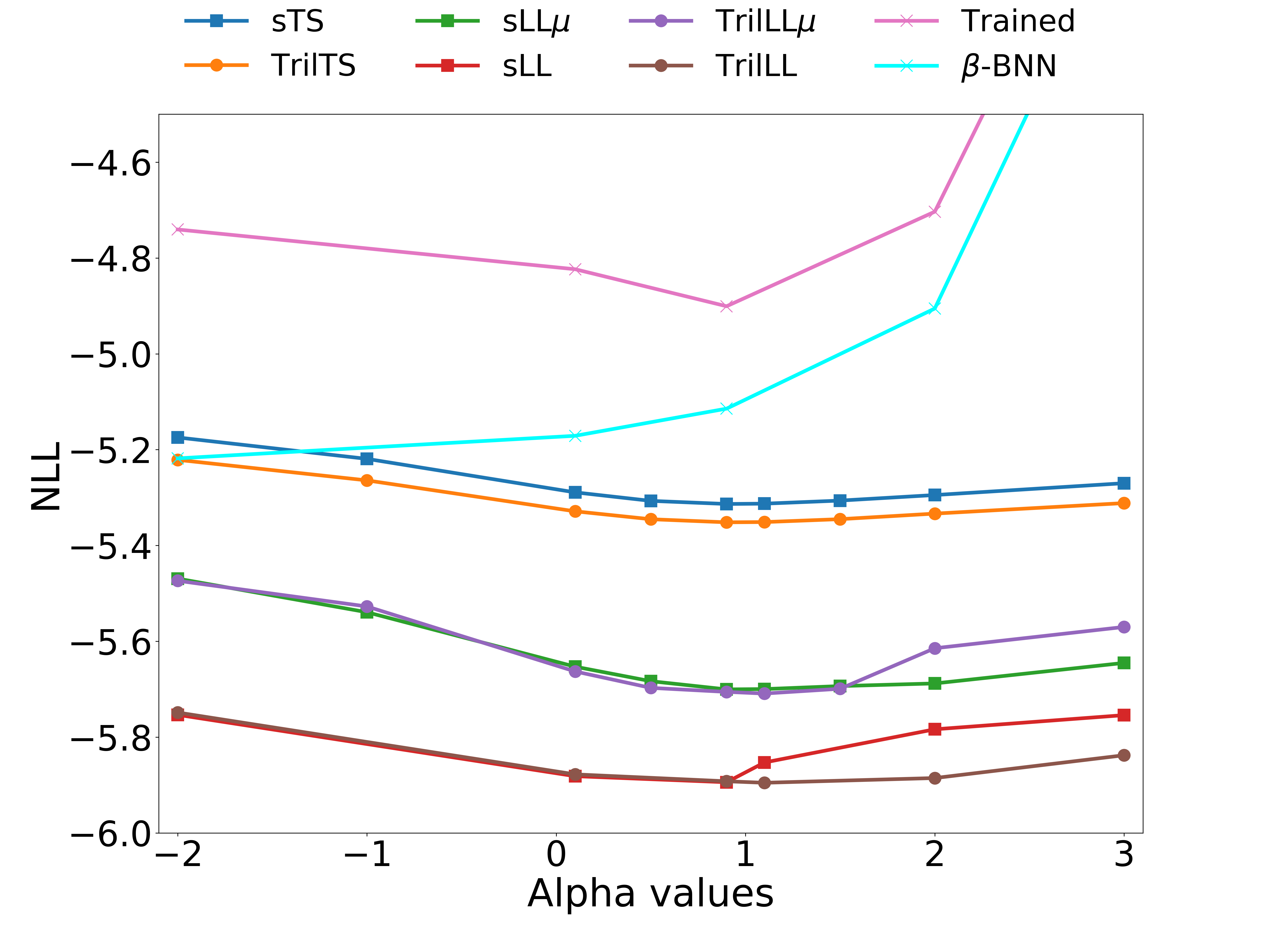}
     \caption{Test NLL of BB-$\alpha$ for both the trained BNN model, and after its post-process calibration. }
     \label{figNLL}
\end{figure}

\begin{figure}[h!]
  \centering
    \includegraphics[width=0.48\textwidth]{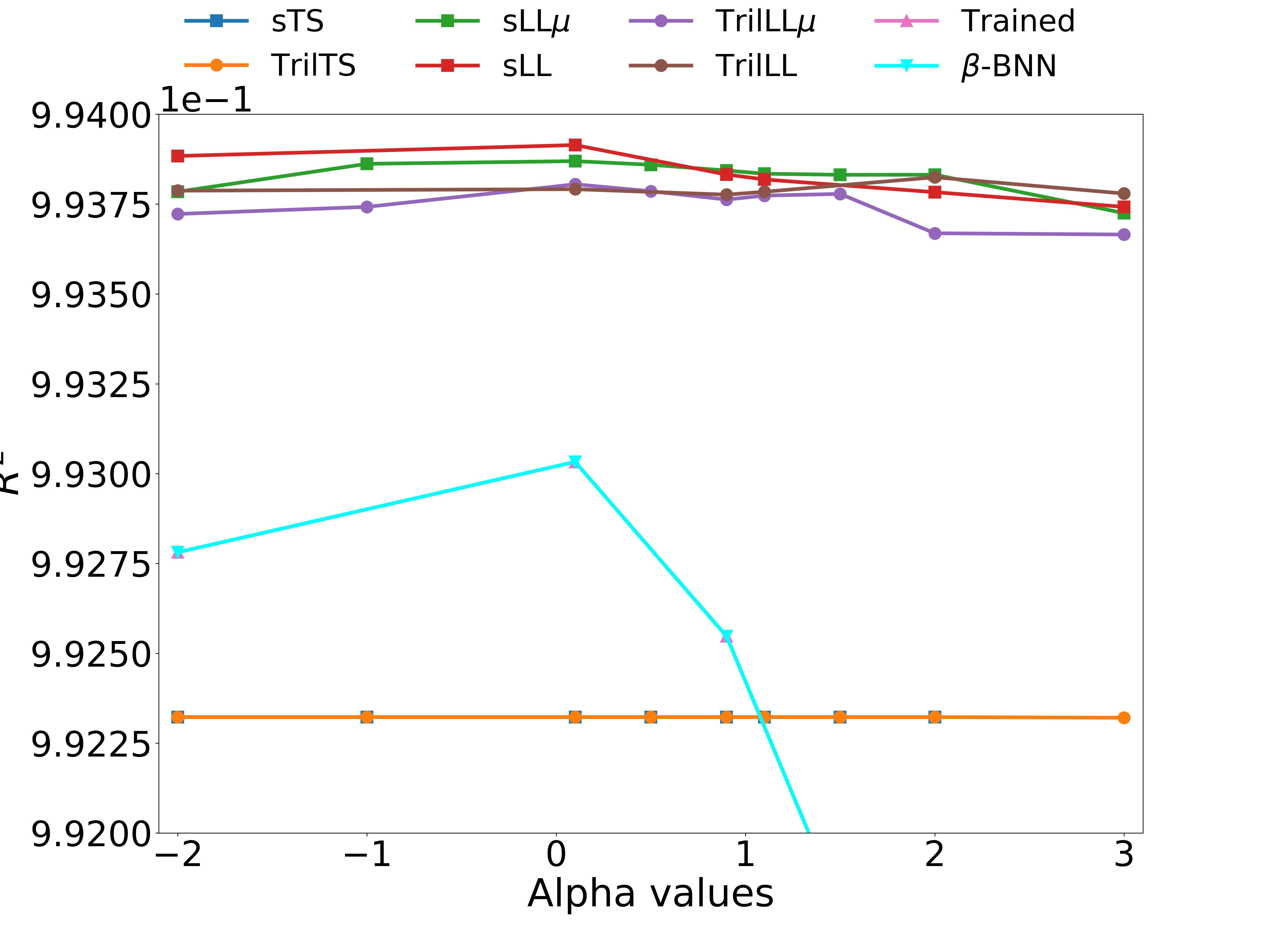}
     \caption{Test $R^2$ of BB-$\alpha$ for both the trained BNN model, and after its post-process calibration.}
     \label{figR2}
\end{figure}

\begin{figure}[h!]
  \centering
    \includegraphics[width=0.5\textwidth]{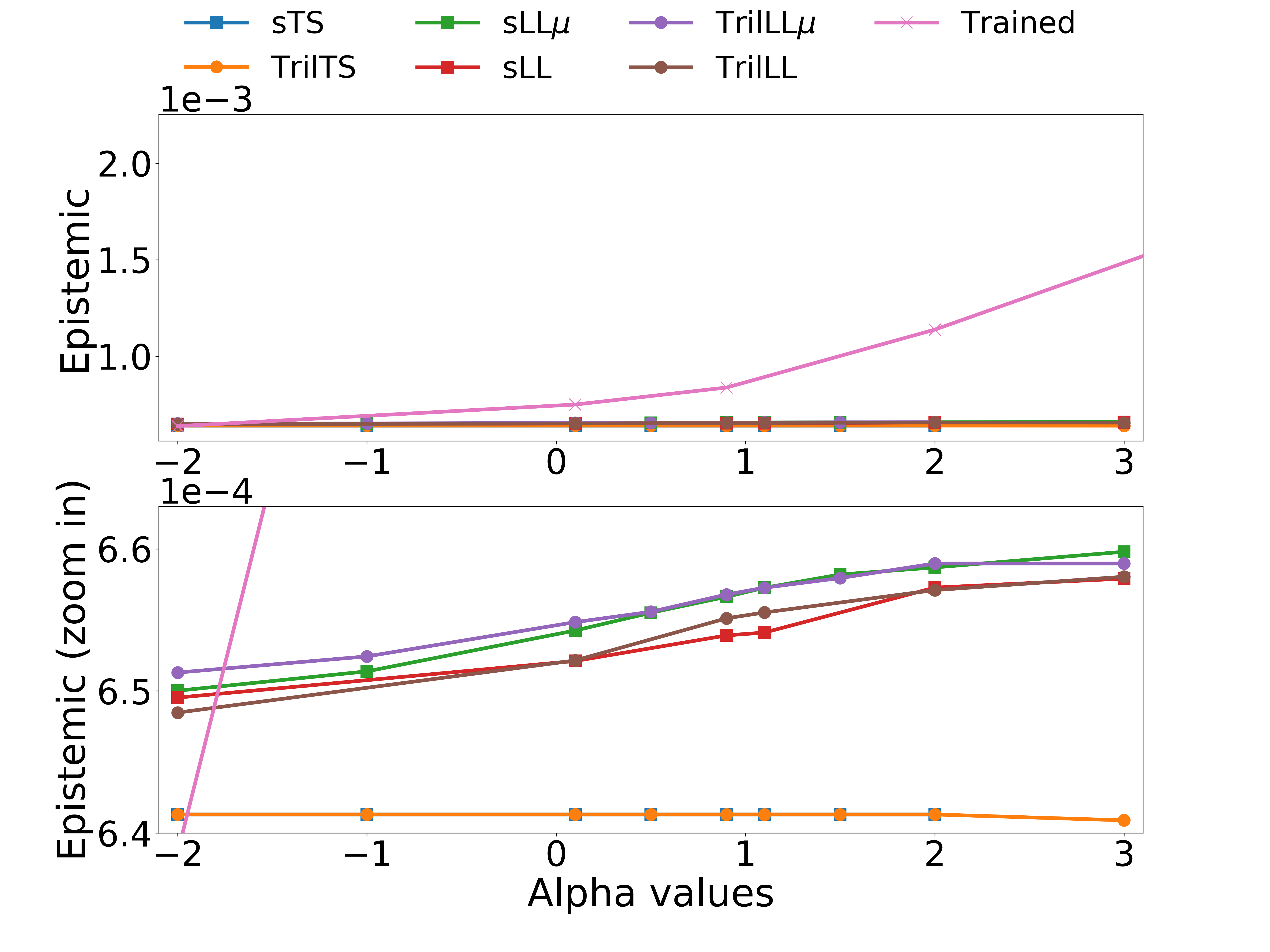}
     \caption{Epistemic uncertainty behavior in  post-process calibration scenario. }
     \label{figepist}
\end{figure}

In Fig.~\ref{figarea} we report calibration experiments with alpha  divergences of the BNN previously trained with standard KL.
This figure shows the area between the miscalibrated and perfect calibrated network, where the horizontal black line corresponds to a perfect calibration.   As previously observed, the BNNs trained with different alpha divergences do not intersect the black line, i.e. they will not produce calibrated uncertainties. In contrast, all the proposed  techniques demonstrate good calibration for some range of alpha around $1$, $\alpha \in [1, 1.5]$. This result is consistent with the test NLL measures  displayed in Fig.~\ref{figNLL}, where the values which comes from the post processing methods outperform the ones obtained by the trained BNNs. 
Moreover, we can observe that LastLayer (LL) usually outperform $\beta$-BNN, and both LastLayermean (LL$\mu$) and Temperature Scaling(TS) in terms of this metric (parameter estimation can be seen in Appendix~\ref{appedB}).  Additionally, the NLL achieves a minimum value  around $\alpha \approx 1$ where the network is adequately calibrated. The coefficient of determination $R^2$ is displayed in Fig.~\ref{figR2}. Note that the ability of BNNs to predict the correct outcomes becomes higher for negative $\alpha$ which turns out to be opposite to the NLL behavior. These results are in agreement with~\cite{2019arXiv190906945R}, where they report that the choice of $\alpha$ depends on the metric we are most interested in, i.e., when $\alpha \approx 1$, the optimization gives more attention to the NLL resulting in a more accurate predictive distribution, while negative $\alpha$'s  focus more on the minimization of the error for  predicting the data.
Finally, Fig~\ref{figepist} displays the effect of $\alpha$ on the epistemic uncertainty. As discussed in the previous section TS alone is not able to affect epistemic uncertainty and thus sTS and TrilTS are flat and overlapping.
Left aside TS, the epistemic increases with the value of $\alpha$ for all methods.
Positive $\alpha$ values tends to cover the entire true posterior, which translate into increasing the variance  and hence the epistemic uncertainty.
Conversely, negatives $\alpha$ produce smaller epistemic values because the approximate posterior favours the local modes in the true posterior implying less variance.
This behaviour of the alpha divergence has been already pointed out in~\cite{minka2005divergence}. 

\section{Conclusions}
In this paper we have presented an extension of temperature scaling method combined with the use of alpha divergences as a set of calibration techniques for BNNs in regression problems, which  improves the performance of the networks both in terms of accurate uncertainties and coefficient of determination $R^2$.
This approach outperforms the use of alpha divergence in training both without and with beta calibration.
Future work will investigate the application of our approach in standard UCI datasets both for regression and classification tasks.

 \section*{Acknowledgements}
 H.J.~Hort\'ua, R.~Volpi, and L.~Malag\`o are supported by the DeepRiemann project, co-funded by the European Regional Development Fund and the Romanian Government through the Competitiveness Operational Programme 2014-2020, Action 1.1.4, project ID P\_37\_714, contract no. 136/27.09.2016.

\bibliography{Paper_ICML_Hortua}
\bibliographystyle{icml2020}

\appendix
\section{Supplemental Materials}
\subsection{Reliability Diagrams}\label{appedA}
Reliability diagrams are  visual representations of model calibration~\cite{Guo:2017:CMN:3305381.3305518}. In our case we generate these diagrams by plotting the predicted interval coverage probabilities  of the test set. The coverage probabilities are defined as the $x\%$ of samples for which the true value of the parameters falls in the $x\%$-confidence intervals. If the model is perfectly calibrated, then the diagram should plot the identity function, and  any deviation from the identity represents miscalibration. For models that  produce an approximately Gaussian joint distribution (higher-order statistical moments are very close to zero),  one can assume that the predictive distribution obey to a multivariate Gaussian distributions whose confidence region can be computed by
\begin{equation}\label{eq:18cal}
\mathcal{C}\geq(\bm{y}-\bm{\hat{y}})^\top \Sigma^{-1}(\bm{y}-\bm{\hat{y}}),
\end{equation}
 which is basically an ellipsoidal confidence set with cove\-rage probability $1-\gamma$. The quantity $\mathcal{C}$ has the Hotelling's T-squared distribution  $T^2_{k,D-k}(1-\gamma)/D$, with $k$ degrees of freedom, being $D$ the number of samples~\cite{hortua2019parameters}.
In  case of unimodal models (not necessarily Gaussian), we can follow the method used in~\cite{PerreaultLevasseur:2017ltk} where we can generate a histogram from binned samples drawn from the  posterior.   Since  this  histogram  is  expected  to  be unimodal, we can compute the interval that contains the (100$\alpha$)\% of the samples around the mode, with $\alpha \in[0,1]$.

\subsection{Parameter model constraints}\label{appedB}
\begin{figure}[h!]
  \centering
    \includegraphics[width=0.52\textwidth]{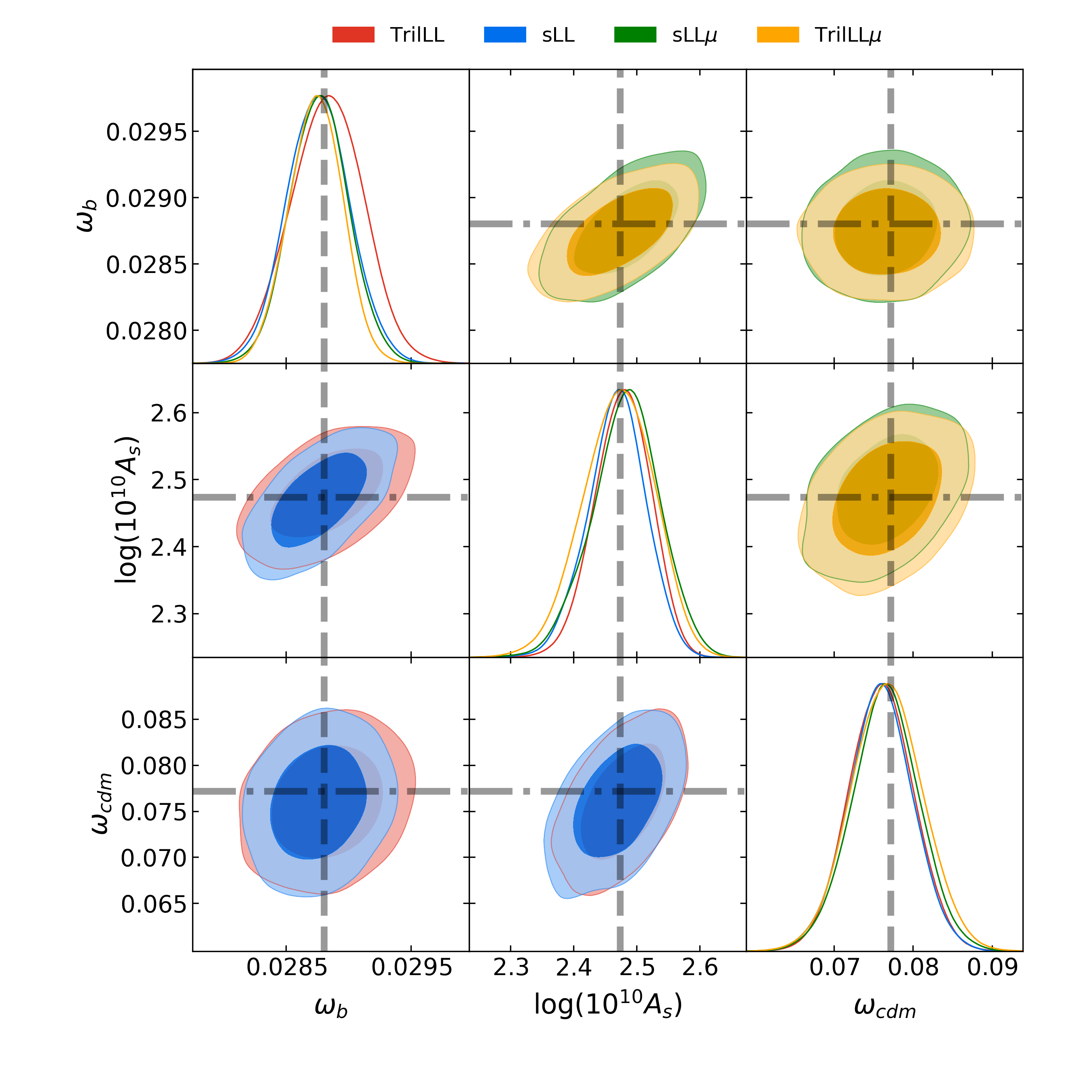}
     \caption{Constraints on parameters of the model from the best calibration approaches.  Contours contain 68\% and 95\% of the probability. The dashed lines represent the true value for an example in the test dataset.}
     \label{figcontourns}
\end{figure}

\subsection{Reliability diagram for a  proposal}\label{appedC}
\begin{figure}[h!]
  \centering
    \includegraphics[width=0.45\textwidth]{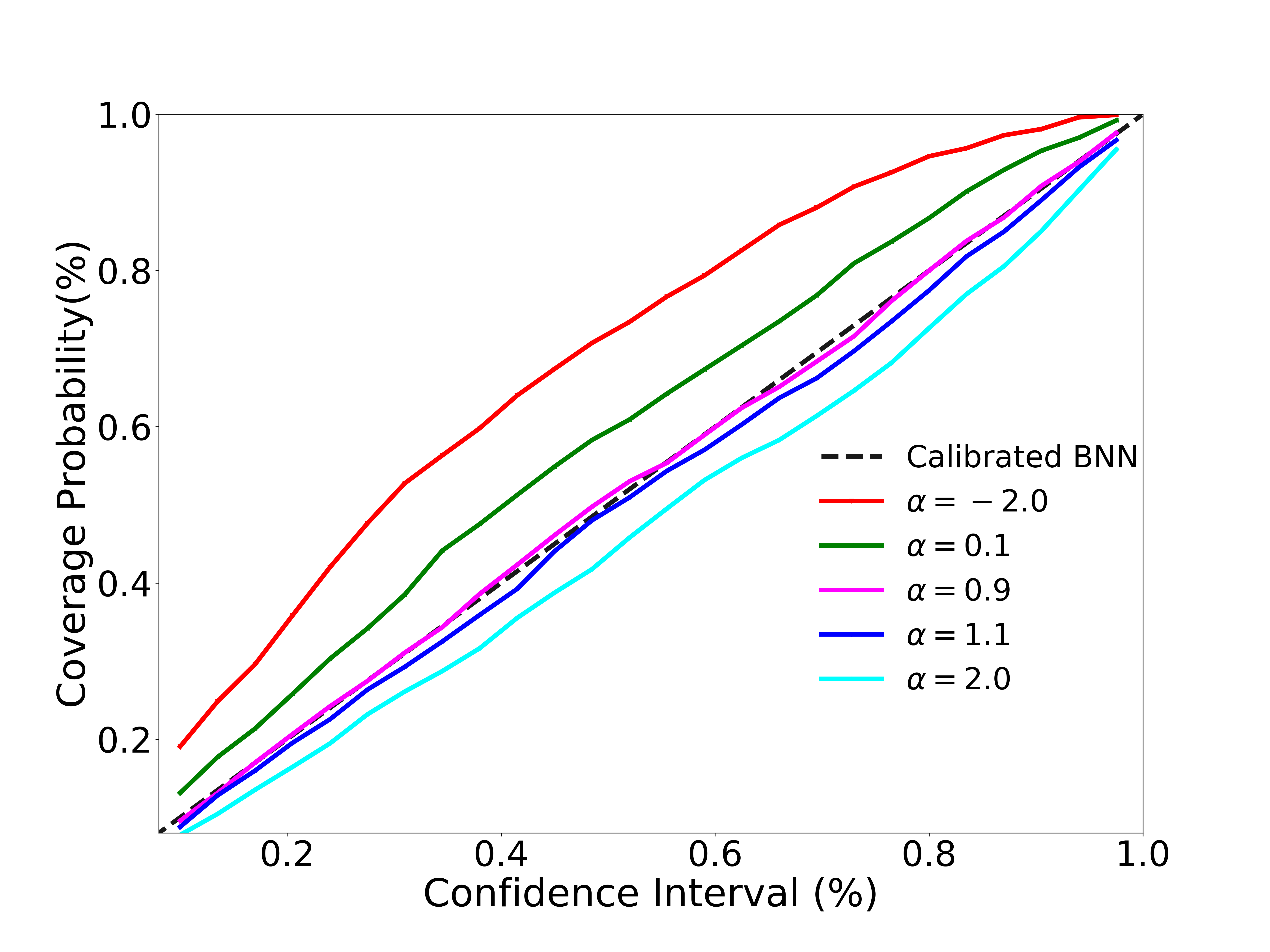}
     \caption{Reliability diagrams for different $\alpha$-values  using the TrilLL approach.}
     \label{figcontourns}
\end{figure}


\end{document}